\pdfoutput=1
\documentclass[10pt,twocolumn,letterpaper]{article}

\usepackage{iccv}
\usepackage{times}
\usepackage{epsfig}
\usepackage{graphicx}
\usepackage{amsmath}
\usepackage{amssymb}

\usepackage{cite}
\usepackage{float}
\usepackage{algorithm}
\usepackage{algorithmicx}
\usepackage{algpseudocode}
\usepackage{mathtools}
\usepackage{url}
\usepackage[table]{xcolor} 

\usepackage{threeparttable,booktabs}
\usepackage{etoolbox}
\appto\TPTnoteSettings{\footnotesize}

\usepackage{arydshln}
\usepackage[skip=0pt]{caption}
\usepackage{subcaption}
\usepackage{makecell}
\usepackage{pifont}
\usepackage{multirow}
\usepackage{booktabs}
\usepackage{chngcntr}

\usepackage{threeparttable}

\usepackage[pagebackref=true,breaklinks=true,letterpaper=true,colorlinks,bookmarks=false]{hyperref}
\iccvfinalcopy 

\def\iccvPaperID{2} 

\ificcvfinal\fi
\begin{document}

\title{Forced Spatial Attention for Driver Foot Activity Classification}

\author{Akshay Rangesh and Mohan M. Trivedi\\
Laboratory for Intelligent \& Safe Automobiles, UC San Diego\\
{\tt\small \{arangesh, mtrivedi\}@ucsd.edu}
}

\maketitle
%
%
%
%


\begin{abstract}
This paper provides a simple solution for reliably solving image classification tasks tied to spatial locations of salient objects in the scene. Unlike conventional image classification approaches that are designed to be invariant to translations of objects in the scene, we focus on tasks where the output classes vary with respect to where an object of interest is situated within an image. To handle this variant of the image classification task, we propose augmenting the standard cross-entropy (classification) loss with a domain dependent Forced Spatial Attention (FSA) loss, which in essence compels the network to attend to specific regions in the image associated with the desired output class. To demonstrate the utility of this loss function, we consider the task of driver foot activity classification - where each activity is strongly correlated with where the driver's foot is in the scene. Training with our proposed loss function results in significantly improved accuracies, better generalization, and robustness against noise, while obviating the need for very large datasets.
\end{abstract}

\section{Introduction}\label{sec:introduction}

Image classification being one of the fundamental tasks in computer vision receives large amounts of research effort, and consequently sees remarkable progress year after year~\cite{szegedy2015going, ioffe2015batch, szegedy2017inception, xie2017aggregated, zoph2018learning, real2018regularized, tan2019efficientnet}. This is true, especially for applications with sufficient training data per class, which is a well understood problem. To ensure better generalization, traditional image classification approaches introduce certain inductive biases, one of which is invariance to spatial translations of objects in images, i.e. the locations of objects of interest in an image does not change the true output class of the image. This is typically enforced by data augmentation schemes like random translations, rotations, crops etc. Even convolution kernels - the basis of most Convolutional Neural Networks (CNNs) are shared across the entire spatial extent of features as a means to learn translation invariant features. In this paper, we are interested in image classification applications where these assumptions do not necessarily hold true. Specifically, we focus on tasks where relative locations of objects in the scene influence the output class of the image. 

\begin{figure}[t]
\begin{center}
\includegraphics[width=\linewidth]{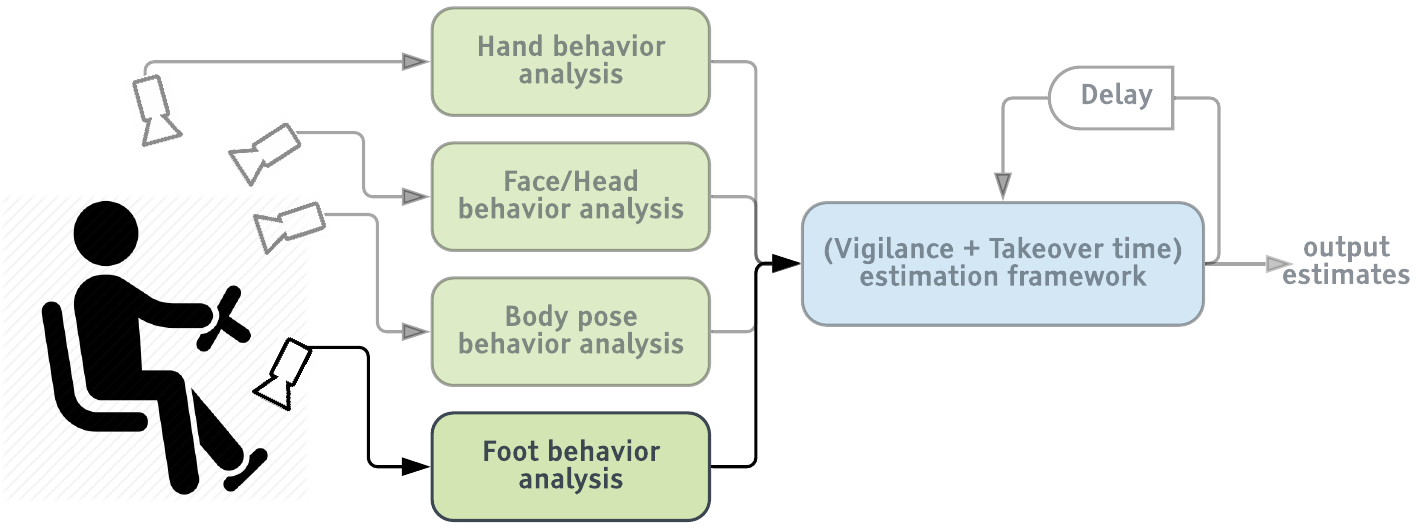}
\end{center}
\caption{Overview of the pipeline to continuously estimate the driver's vigilance and readiness to takeover control from a semi-autonomous driving agent. We highlight the parts relevant to this study (foot behavior analysis) in bold. To ensure smooth control transitions, the final estimation framework, and thus each individual analysis block that feeds into it needs to be reliable and robust under a variety of conditions.}
\label{fig:motivation}
\end{figure}

Many real world examples of such tasks can be found in the surveillance domain. For example, consider the scenario where we would like identify when an unauthorized person is in close proximity to a stationary object like a car/door/safe etc. If this were set up as an image classification problem, the desired output class would vary based on where the unauthorized person is in the image i.e. if the person is very close to the stationary object and exhibiting unusual behavior, then trigger an alarm; else do not. In this study, we try to solve a similar problem from the automotive domain. In particular, we wish to design a very simple and reliable system to classify the foot activity of drivers in cars. This problem is comprised of 5 classes of interest, namely - \textit{away from pedals}, \textit{hovering over accelerator}, \textit{hovering over break}, \textit{on accelerator}, and \textit{on break}. As can be inferred from the individual class identities, the desired output changes based on where the driver's foot is in the image. We chose these classes as they are good indicators of a driver's preparatory motion, and are also strongly tied to the time it takes for a driver completely regain control of the car from an autonomous agent~\cite{rangesh2018exploring, deo2018looking} - also known as the \textit{takeover time}. Figure~\ref{fig:motivation} depicts the goal of this study and its role in solving the bigger problem of driver vigilance and takeover time estimation.

Before describing our approach, we would also like to address some straightforward ways in which one could potentially solve such problems. One obvious way to encode spatial information in predictions is to use a fully connected (FC) output layer. This however comes at a huge cost of computation, storage, and possibly generalization. Introducing an FC layer would also increase the data requirements considerably, something that is not available in many applications. Another way to approach these problems is to split the task into specialized portions, leading to better generalization and interpretability~\cite{gulccehre2016knowledge}. For instance, you could have one algorithmic block dedicated to detecting all objects of interest in an image, followed by a second block that would reason over their spatial locations. The major drawback of such approaches is the requirement of ground truth object locations in the image for training the individual blocks. Once again, this is quite expensive to obtain and is not available in many applications of interest. Our proposed approach attempts to reliably solve this class of problems without introducing any of the aforementioned drawbacks.

Our main contributions in this work can be summarized as follows - 1) We propose a simple procedure to modify the training of CNNs that make use of Class Activation Maps (CAMs)~\cite{zhou2016learning} so as to introduce spatial and domain knowledge related to the task at hand 2) To this end, we propose a new Forced Spatial Attention (FSA) loss that compels the network to attend to specific regions in the image based on the true output class. 3) Finally, we carry out qualitative and quantitative comparisons with standard image classification approaches to illustrate the advantages of our approach using the task of driver foot activity classification.

\section{Related Research}

\noindent \textbf{Driver foot activity research:} Tran et al. conducted some of the earliest research on modeling foot activity inside cars for driver safety applications. In~\cite{tran2011pedal, tran2012modeling}, they track the driver's foot using optical flow, while maintaining the current state of foot activity using a custom Hidden Markov Model (HMM) comprising of seven states. Maximizing over conditional state probabilities then produces an estimate of the most likely foot activity at any given time step. This system was intended as a solution to identify and prevent pedal misapplications, a common cause for accidents at the time. More recently, Wu et al.~\cite{wu2017foot} propose a more holistic system comprising of features obtained from visual, cognitive, anthropometric and driver specific data. They use two models - a random forest algorithm was used to predict the likelihood of various pedal application types, and a multinomial logit model was used to examine the impact of prior foot movements on an incorrect foot placement. Although these resulted in high classification errors, the authors were able to identify features important for identifying and preventing pedal misapplications. In their following study~\cite{wu2018evaluating}, the authors analyze foot trajectories from a driving simulator study, and use Functional Principal Component Analysis (FPCA) to detect unique patterns associated with early foot movements that might indicate pedal errors. Inspired by previous work, the Zeng et al.~\cite{zeng2017stochastic} also incorporated vehicle and road information by looking outside the vehicle to model driver pedal behavior using an Input-Output HMM (IOHMM). Unlike most other methods that make use of potentially privacy limiting video sensors, the authors in~\cite{frank2019robust} use capacitive proximity sensors to recognize four different foot gestures. 

Driver foot activity has also been an area of interest for many human factors studies. Recent examples include~\cite{mcgehee2016wagging}, where the authors collect and reduce naturalistic driving data to identify and understand problematic behaviors like pressing the wrong pedal, pressing both pedals, incorrect trajectories, misses, slips, and back-pedal hooks etc. Elsewhere, Wang et al.~\cite{wang2017bipedal} conduct a simulator based study to compare unipedal (using the right foot to control the accelerator and the brake pedal) and bipedal (using the right foot to control the accelerator and the left foot to control the brake pedal) behavior among drivers. They found the throttle reaction time to be faster in the unipedal scenario, whereas brake reaction time, stopping time, and stopping distance showed a bipedal advantage. For a more detailed and historical perspective on driver (and human) foot behavior and related studies, we refer the reader to~\cite{velloso2015feet, ohn2016looking, doshi2011tactical, leo2017computer, farinella2016special}.

\smallskip
\noindent \textbf{Class Activation Maps (CAMs):} In this study, we manipulate CAMs by \textit{forcing} them to activate only at certain predefined regions depending on the output class. CAMs originated from weakly-supervised classification research~\cite{zhou2016learning}, where the authors demonstrated that using a Global Average Pooling (GAP) operation instead of an output FC layers resulted in per-class feature maps that loosely localize objects of interest. This offered additional benefits such as relatively better interpretability and reduced model size. More recently, several studies have tried to improve the localization in CAMs in the weakly supervised regime. Singh et al.~\cite{singh2017hide} improve the localization in CAMs by randomly hiding patches in the input image, thereby forcing the network to pay attention to other relevant parts that contribute to an accurate classification. Other popular methods~\cite{wei2017object, kim2017two, li2018tell} typically contain multiple stages of the same network. The CAMs from the first stage are used to mask out the inputs/features to the second stage, thereby forcing the network to pay attention to other salient parts of an image. This results in a more complete coverage of parts relevant to the true class of an image.

\begin{figure*}[t]
\begin{center}
\includegraphics[width=\linewidth]{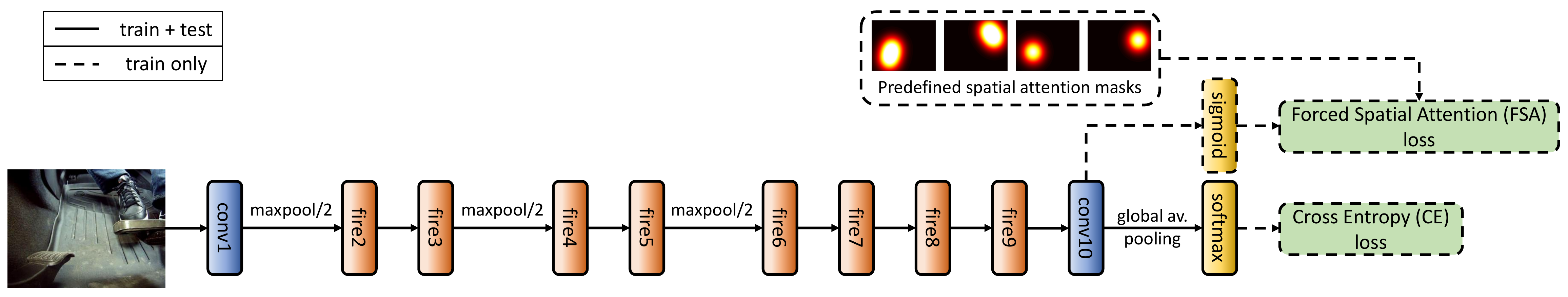}
\end{center}
\caption{Proposed network architecture for training and inference. The network is based on the Squeezenet v1.1 architecture~\cite{SqueezeNet} with an additional training-only output branch used to \textit{force} the network's spatial attention.}
\label{fig:network}
\end{figure*}

\section{Methodology}

\subsection{Network Architecture}

Our primary focus in this study is to propose a general procedure for training CNNs for image classification in a setting where the output classes are tied to domain dependent spatial locations of activity. Although any CNN architecture could be chosen, we decide to work with the Squeezenet v1.1 architecture~\cite{SqueezeNet} for the following reasons: the Squeezenet model is extremely lightweight and therefore less data-hungry, while still retaining sufficient representation power. The model also makes use of CAMs instead of FC layers, thereby making it naturally amenable to the proposed FSA loss that we apply to the normalized CAMs. It must however be noted that models with FC layers can also be made compatible with our procedure by using Gradient-weighted Class Activation Maps (Grad-CAMs)~\cite{selvaraju2017grad}. Finally, using a lightweight architecture like Squeezenet is extremely useful deployment in the real world, where power and computational efficiency are critical. 

Most of our experiments begin with a Squeezenet v1.1 model pretrained on Imagenet. During training, we augment the existing architecture with a Forced Spatial Attention (FSA) head that branches off from the existing \textit{conv10} layer that produces the CAMs, before the global average pooling operation (GAP) is applied. This modification is illustrated in Figure~\ref{fig:network}. The FSA head takes as input the CAMs, then normalizes them to $[0, 1]$ through a sigmoid operation. These normalized CAMs along with predefined, domain dependent spatial masks are then used to compute the FSA loss which is backpropagated throughout the network along with the conventional cross entropy (classification) loss. The FSA head and the corresponding FSA loss are used only during training, as a means to inject domain specific spatial knowledge into the network. Once trained, the FSA head is removed and the architecture reverts to its original form. 

\subsection{Forced Spatial Attention}

\begin{figure*}[t]
\begin{center}
\includegraphics[width=0.85\linewidth]{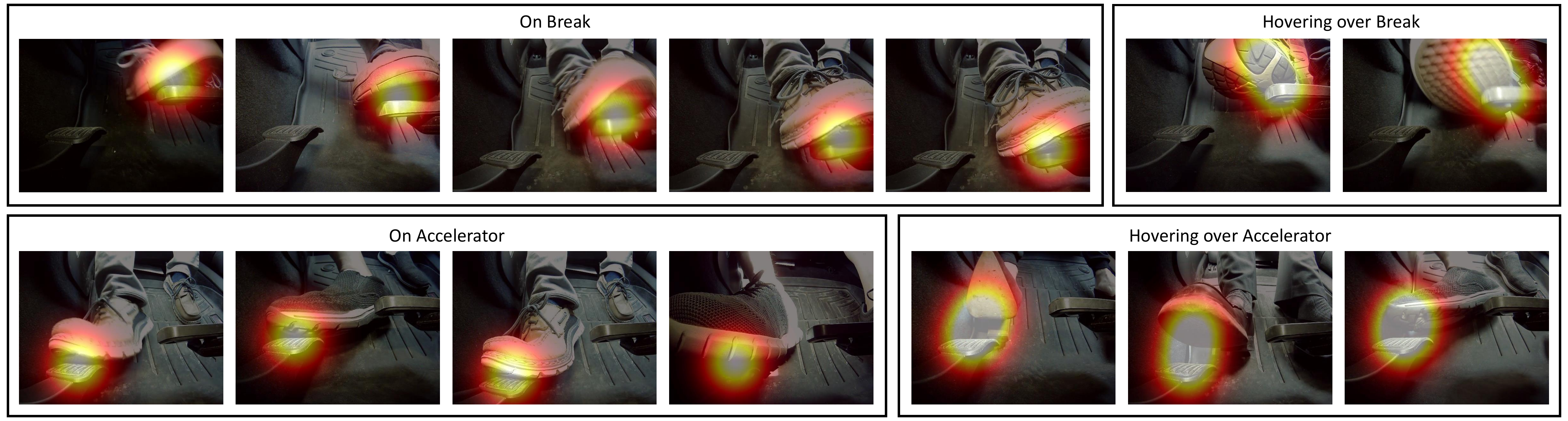}
\end{center}
\caption{Predefined spatial attention masks for each class overlaid on an exemplar input images from the class. Classes are associated with multiple attention masks to account for different foot positions during activities, and slight camera movements. The class \textit{away from pedals} is not associated with a spatial attention mask and has been omitted above.}
\label{fig:attention_masks}
\end{figure*}

Class Activation Maps (CAMs) are generally used as a means to provide visual reasoning for observed network outputs, i.e. to understand which regions a network attended to, while producing the observed output. Conversely, if one knows which spatial locations the network must attend to for a desired output class, this can be used as a supervisory signal to train the network. If done correctly, this should reduce overfitting and improve generalization, as the network is \textit{forced} to attend to relevant regions only, while ignoring extraneous sources of information. This is the goal of our proposed FSA loss. We explain this loss more concretely in the context of our desired application, i.e. driver foot activity classification.

The goal of our driver foot activity classification task is to predict one of five activity classes: \textit{away from pedals}, \textit{hovering over accelerator}, \textit{hovering over break}, \textit{on accelerator}, and \textit{on break}, using images from a camera observing the driver's foot inside a vehicle cabin. Examples of these images are provided in Figure~\ref{fig:attention_masks}. The next step in our procedure is to create spatial attention masks for some/all output classes. The key idea is to create spatial attention masks with \textit{peaks} at regions depicting the activity corresponding to the output class. Examples of these predefined attention masks for various images and different classes are illustrated in Figure~\ref{fig:attention_masks}. Note that the \textit{Away from pedals} class is not associated with any attention maps because it is not tied to any spatial location by definition. On the other hand, certain classes are associated with multiple spatial locations due to slight changes in camera perspective, and also because of the very nature of the activity. For example, the activity \textit{On break} could be associated with different attention masks depending on how far the break pedal is pushed (see Figure~\ref{fig:attention_masks}). One issue with having multiple attention masks per class is that we do not know which mask is to be used for a given training image. We address this issue using a two stage training approach described below. 

Let $A^C$ denote the CAM and $\mathcal{H}^C = \{H_1^C, H_2^C,\cdots, H_{N_C}^C\}$ denote the set of predefined spatial attention masks for class $C$. Note that the number of spatial attention masks $N_C$ could be different for each class $C$. Our classes range from $C=1, 2, \cdots, 5$ to indicate the five possible output classes. As mentioned earlier, we first apply a pixelwise sigmoid transformation to the CAMs to normalize them to $[0, 1]$:
\begin{equation}
T(A^C) = \frac{1}{1+\exp(-A^C)}.
\end{equation}

Next, to resolve the ambiguities arising from having multiple predefined attention maps per class, we use a two stage training procedure - each associated with a different FSA loss. In the first stage, we force the network to attend to all possible regions of interest per class. This is achieved through the loss function:
\begin{equation}
\begin{split}
L_{FSA}^{stage-1} = \big(T(A^{C^*}) - \max(\mathcal{H}^{C^*})\big)^2 +\\ \lambda_{FSA}^{reg} \sum_{\substack{C=1 \\C \neq C^*}}^{5} \text{mean}\big(T(A^{C^*}) \odot T(A^C)\big),
\end{split}
\end{equation}
where $C^*$ denotes the ground truth class for a given input image, $\max(\mathcal{H}^{C^*})$ denotes the pixelwise maximum operation applied to all transformed attention maps of the true class $C^*$, and $\odot$ denotes the Hadamard product between two matrices. We note that the first term of the FSA loss is simply the MSE loss between the ground truth CAM and the pixelwise maximum of all predefined attention masks belonging to the same class. The second term is a regularizer term to encourage independence between CAMS. We observe that omitting the second term leads to \textit{activation leakage}, where CAMs for other classes have high activations in spatial locations corresponding to the ground truth class.
The total loss for the network in stage-1 of training is thus given by
\begin{equation}\label{eq:stage-1}
L^{stage-1} = L_{CE} + \lambda_{FSA} L_{FSA}^{stage-1},
\end{equation}
where $L_{CE}$ denotes the standard cross entropy classification loss.

In stage-1 of training, the network is forced to attend to all possible regions of interest for a specific class. In stage-2 of training, we would like the network to \textit{contract} its attention to the region pertinent to the input image. With this in mind, the FSA loss for stage-2 is defined as:
\begin{equation}
\begin{split}
L_{FSA}^{stage-2} = \min_{i}\bigg(\big(T(A^{C^*}) - H_i^{C^*})\big)^2\bigg) +\\ \lambda_{FSA}^{reg} \sum_{\substack{C=1 \\C \neq C^*}}^{5} \text{mean}\big(T(A^{C^*}) \odot T(A^C)\big),
\end{split}
\end{equation}
where we modify only the first term of the FSA loss. Specifically, we only apply an MSE loss between the ground truth CAM and the predefined attention mask that is most similar in an L2 sense. The reasoning behind this is to make the network choose attention masks that retain features that are most discriminative for each input image. 
As before, the total loss for the network in stage-2 of training is given by
\begin{equation}\label{eq:stage-2}
L^{stage-2} = L_{CE} + \lambda_{FSA} L_{FSA}^{stage-2}.
\end{equation}
We demonstrate through our experiments that such a two stage loss results in the network learning to choose the correct attention mask without explicit supervision.

\subsection{Implementation Details}

\begin{figure}[t] 
   \captionsetup[subfigure]{justification=centering}
   \centering
   \begin{subfigure}[t]{0.48\linewidth}
      \centering
      \includegraphics[width=\linewidth]{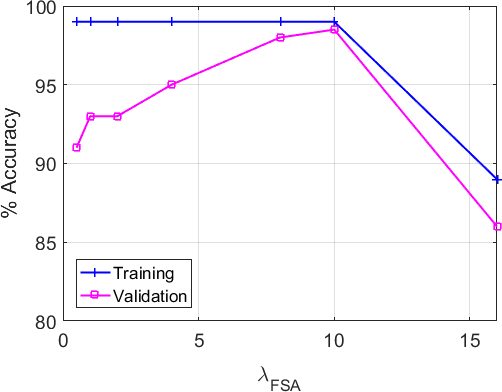}
      \caption{}
   \end{subfigure}%
~     
   \begin{subfigure}[t]{0.48\linewidth}
      \centering
      \includegraphics[width=\linewidth]{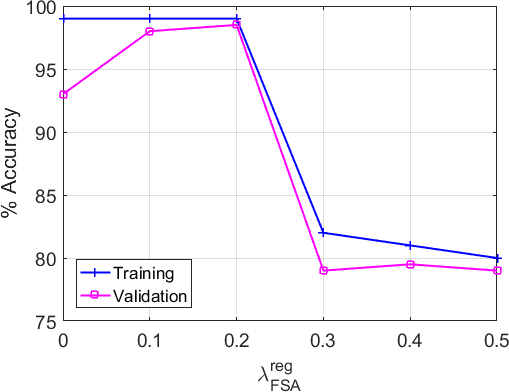}
      \caption{}
   \end{subfigure}
   \caption{Plot of training and validation accuracies for different values of hyperparameter (a)$\lambda_{FSA}$ and (b)$\lambda_{FSA}^{reg}$.}
   \label{fig:hyperparam}
\end{figure}

To create the class specific attention masks $\mathcal{H}^C$, we first collected a set of representative images for each class. These images were chosen to represent the different regions of activity within a given class. Next, we created the various attention masks by manually overlaying a 2D Gaussian peak with suitable variance over each image. For certain classes such as \textit{hovering over accelerator}, we placed two Gaussian peaks in close locality to cover the larger spatial extent of such activities. The resulting attention masks for each class are depicted in Figure~\ref{fig:attention_masks}.   

For our classification model, we initialize the Squeezenet v1.1 model with Imagenet pretrained weights. The training is carried out in two stages for a total of 30 epochs. Standard mini batch Stochastic Gradient Descent (SGD) with a batch size of $64$ is used to train the network. We use a learning rate of $0.0005$ with a momentum equal to $0.9$, and a weight decay term to reduce model complexity. 

\begin{figure}[t]
\begin{center}
\includegraphics[width=0.9\linewidth]{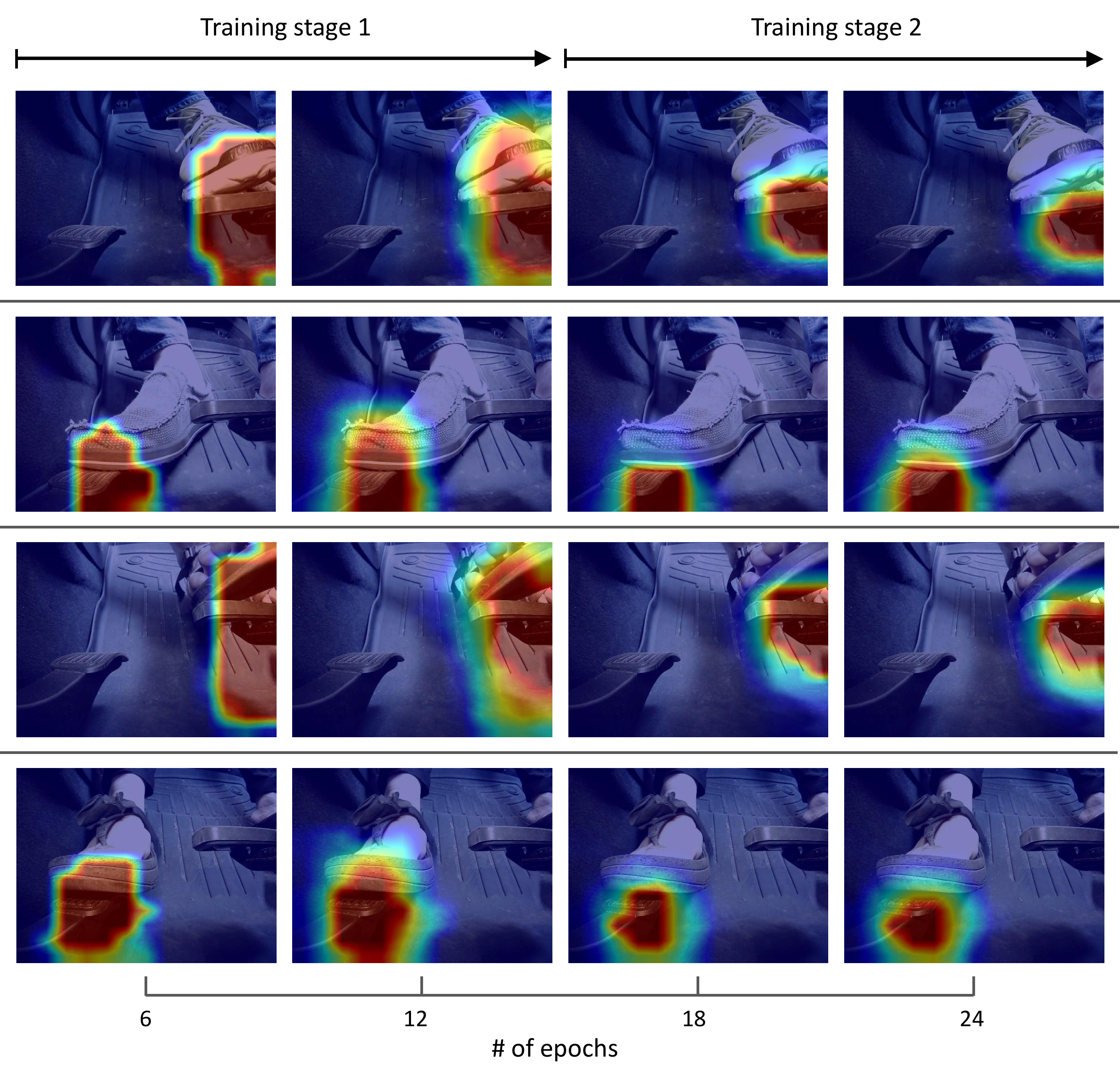}
\end{center}
\caption{Class Activation Maps (CAMs) for the correct output class as a function of the number of training epochs. Each row is a different example.}
\label{fig:attention_masks_epochs}
\end{figure}

The network is trained for the first 15 epochs using the $L^{stage-1}$ loss (Eq.~\ref{eq:stage-1}), and then using the $L^{stage-2}$ loss for the remaining epochs. The hyperparameters $\lambda_{FSA}$ and $\lambda_{FSA}^{reg}$ are determined through extensive cross-validation, the results of which are shown in Figure~\ref{fig:hyperparam}. Our final choices for hyperparameters $\lambda_{FSA}$ and $\lambda_{FSA}^{reg}$ were $10$ and $0.2$ respectively.
The qualitative effect of our two stage training approach is illustrated in Figure~\ref{fig:attention_masks_epochs} for further clarity. In the depicted examples from the training and validation sets, we observe that during the first stage of training, the network learns to attend to large regions corresponding to various possible regions of activity, while the region of attention gradually contracts to the specific region of activity corresponding to the given input image in the second stage of training. In particular, we observe that the attention contracts to the location where the foot hits the pedal, for different locations of the foot and pedal.

\section{Experimental Evaluation}

\subsection{Dataset}

To train and evaluate our proposed model and its variants, we collect a diverse dataset of images capturing driver foot activities. This data was collected during naturalistic drives, with many different drivers as subjects. Details of our complete dataset and the \textit{train}, \textit{validation}, and \textit{test} splits are listed in Table~\ref{table:dataset}. In particular, we ensure that no subjects overlap between the three splits so as to test the cross-subject generalization of our models. We also try our best to keep the class distributions similar across the three splits.

\begin{table}[ht]
\centering
\tabcolsep=0.09cm
\caption{Details of the train-val-test split used for the experiments.}
\resizebox{0.6\columnwidth}{!}{%
 \begin{tabular}{ c | c | c } 
 \thead{Split} & \thead{Number of\\unique drivers} & \thead{Number of images}\\
 \hline
 \thead{Train} & $7$ & $19,385$\\ 
 \thead{Validation} & $1$ & $1,867$\\
 \thead{Test} & $3$ & $7,698$\\
 \end{tabular}
 }
 \label{table:dataset}
\end{table}

\subsection{Results}

\begin{table}[t]
\centering
\tabcolsep=0.09cm
\caption{Classification accuracies for different model variants on the test split.}
\begin{threeparttable}
\resizebox{0.8\columnwidth}{!}{%
 \begin{tabular}{| c | c | c |}
 \hline
 \thead{Model} & \thead{Loss} & \thead{Accuracy (\%)}\\
 \hline\hline
 \thead{SqueezeNet v1.1} & \thead{CE} & $85.99$\\
 \hline
 \thead{SqueezeNet v1.1} & \thead{CE+MSE} & $89.67$\\
 \hline
 \thead{SqueezeNet v1.1} & \thead{CE +\\FSA (stage 1 only)} & $92.30$\\
 \hline
 \thead{SqueezeNet v1.1} & \thead{CE +\\FSA (stage 2 only)} & $85.49$\\
 \hline
 \thead{SqueezeNet v1.1} & \thead{CE +\\FSA (both stages)} & $\mathbf{97.49}$\\
 \hline
 \thead{SqueezeNet v1.1 w/ FC output layer} & \thead{CE} & $63.31$\\
 \hline
 \thead{AlexNet v1.1 w/ FC output layer} & \thead{CE} & $60.99$\\
 \hline
 \thead{VGG16 v1.1 w/ FC output layer} & \thead{CE} & $67.03$\\
 \hline
 \end{tabular}
 }
  \begin{tablenotes}
 \item CE: Cross Entropy loss
 \item MSE: Mean Squared Error loss
 \item FSA: Forced Spatial Attention loss
 \end{tablenotes}
\end{threeparttable}
 \label{table:results}
\end{table}

\begin{figure*}[t]
   \captionsetup[subfigure]{justification=centering}
   \centering
   \begin{subfigure}[t]{0.3\linewidth}
      \centering
      \includegraphics[width=\linewidth]{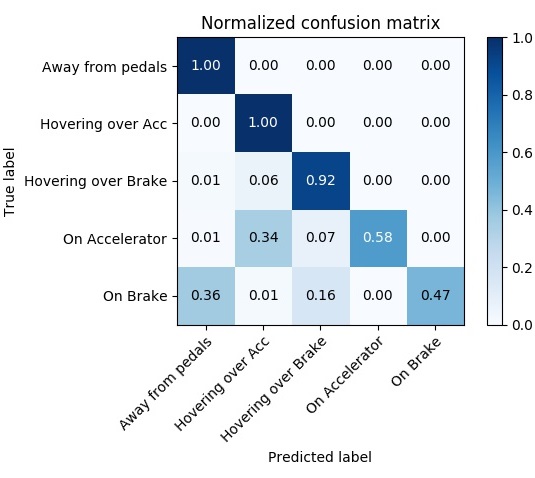}
      \caption{CE loss}
   \end{subfigure}%
~     
   \begin{subfigure}[t]{0.3\linewidth}
      \centering
      \includegraphics[width=\linewidth]{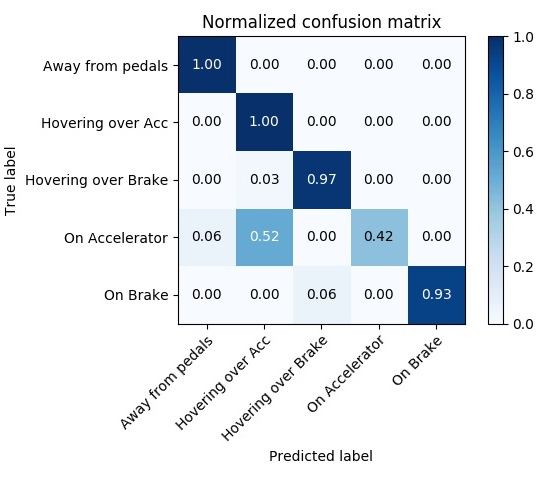}
      \caption{(CE + MSE) loss}
   \end{subfigure}%
~
   \begin{subfigure}[t]{0.3\linewidth}
      \centering
      \includegraphics[width=\linewidth]{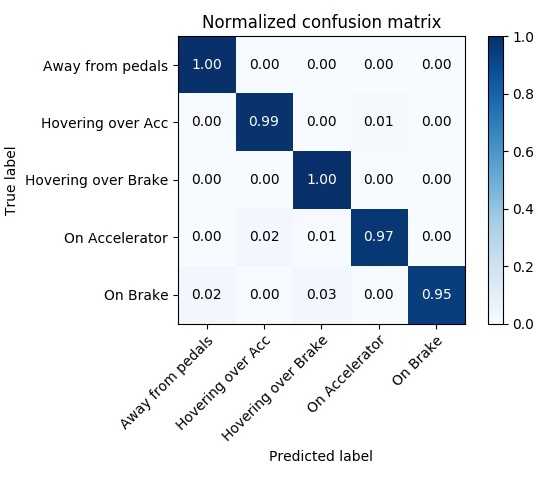}
      \caption{(CE + FSA) loss}
   \end{subfigure}
   \caption{Confusion matrices on the test split for networks trained using different losses.}
   \label{fig:confusion}
\end{figure*}

We first compare the overall classification accuracies of different variants of our Squeezenet v1.1 model on the test split (see Table~\ref{table:results}). All variants of Squeezenet v1.1 were initialized with pretrained Imagenet weights before training. First, we have the model trained only using the standard cross entropy classification loss. This model produces a reasonable accuracy of $85.99\%$ and provides a strong baseline to compare our proposed approach against. Next, we compare different versions of our model that make use of the predefined attention masks during training, but differ in the losses they use to force spatial attention. It is observed that simply incorporating domain specific spatial knowledge leads to an improvement in overall accuracy, irrespective of the specific choice of the loss function. Adding a simple MSE loss (i.e. using only the first term from the loss defined in Eq.~\ref{eq:stage-2}) between the CAMs and their corresponding attention masks leads to a modest improvement over the baseline. We also observe that using either one of the two stages of the FSA loss also improves the overall accuracy, but not as much as when they are used in conjunction over two stages. Our proposed two stage FSA loss leads to the best overall accuracy of $97.49\%$- a significant improvement over the baseline. Finally, we also provide accuracies for Squeezenet v1.1, AlexNet~\cite{krizhevsky2012imagenet}, and VGG16~\cite{simonyan2014very} with output FC layers. Even though an FC layer by nature can produce location specific features, we observe that the large size of the models and limited size of the dataset make it a bad fit for the task at hand.

\begin{figure*}[t]
\begin{center}
\includegraphics[width=0.85\linewidth]{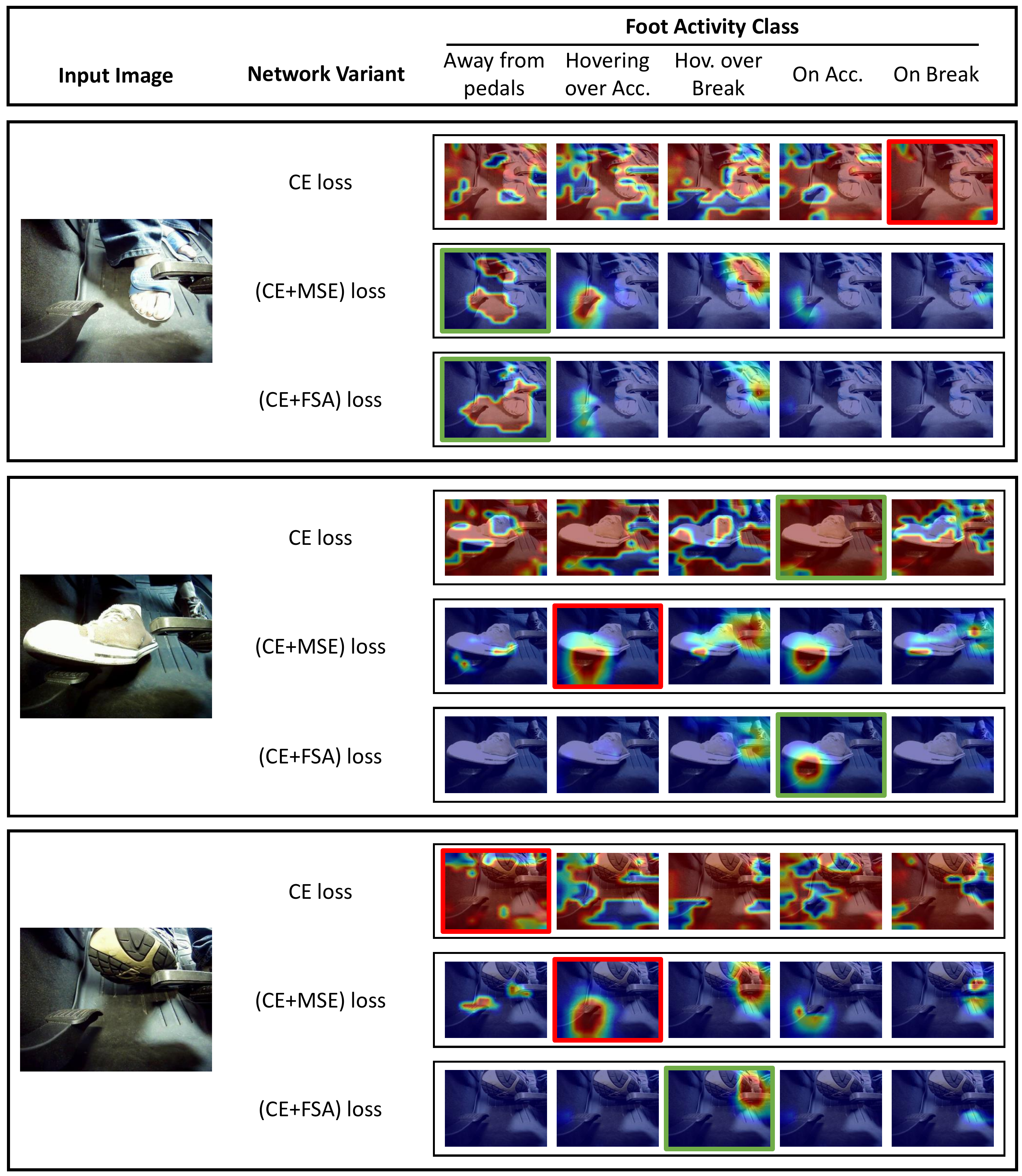}
\end{center}
\caption{Class Activation Maps (CAMs) resulting from networks trained with different loss functions. The three major rows correspond to three different input images. The green boxes shows the ground truth class labels while the red boxes shows if the network made an incorrect prediction.}
\label{fig:heatmaps}
\end{figure*}

We can also gather some insights about the performance of each variant by looking at both their confusion matrices on the test split (Figure~\ref{fig:confusion}) and their CAMs for different input images (Figure~\ref{fig:heatmaps}). Although the baseline model results in a reasonable overall accuracy, it fails to learn the true concept of each class and overfits to background information. This is illustrated by its confusion between classes that are very different to one another and its mostly uniform CAMs. Next, we observe that incorporating domain specific spatial information using predefined attention masks and an MSE loss makes the model better and more robust, with much more informative CAMs. However, we can also see activation leakage between classes (CAMs with high activations in the same region), resulting in confusion between similar classes. Finally, we see that adding a regularizing term as in the two stage FSA loss resolves these issues. It not only reduces the confusion between similar classes, but also produces more confident outputs as illustrated by the corresponding CAMs. 

The failure cases we generally observe are at the boundaries of the \textit{hovering over} $\rule{0.7cm}{0.15mm}$ and the \textit{on} $\rule{0.7cm}{0.15mm}$ classes, especially when the foot is hovering very close to one of the pedals. We find this to be acceptable because of the relative difficulty that humans have in confidently labelling these examples.

\section{Concluding Remarks}
In this study, we introduce a simple approach to solve image classification tasks where the output classes are tied to relative spatial locations of objects in the image. We do so by augmenting the standard classification loss with a Forced Spatial Attention (FSA) loss that compels the network to attend to specific regions in the image associated to the desired output class. The FSA loss function provides a convenient way to incorporate spatial priors that are known for a certain task, thereby improving robustness and generalization without requiring additional labels. The benefits of our approach are demonstrated for the driver foot activity classification task, where we improve the baseline accuracy by approximately 13\% without modifying the network architecture. Such an improvement is especially valuable for ensuring the robustness and reliability of downstream safety critical tasks such as driver vigilance and takeover time estimation. 

\section{Acknowledgments}
We gratefully acknowledge our sponsor Toyota CSRC for their continued support. We would also like to thank our collaborators for helping us collect diverse real-world data.

{\small
\bibliographystyle{IEEEtran}
\bibliography{ref}
}

\end{document}